\documentclass[conference]{IEEEtran}
\IEEEoverridecommandlockouts

\usepackage{graphicx}
\usepackage{amsmath}
\usepackage{amssymb}
\usepackage{booktabs}
\usepackage{multirow}
\usepackage{footmisc}
\usepackage[hyphens]{url}

\usepackage{hyperref}
\hypersetup{
colorlinks=true,
linkcolor=blue,
citecolor=blue,
urlcolor=blue,
}

\begin{document}

\title{Filter-based Discriminative Autoencoders for Children Speech Recognition}

\author{
Chiang-Lin Tai$^{1}$, Hung-Shin Lee$^{1}$, Yu Tsao$^{2}$, and Hsin-Min Wang$^1$\\
$^1$\textit{Institute of Information Science, Academia Sinica}\\
$^2$\textit{Research Center for Information Technology Innovation, Academia Sinica}\\
\{taijohnny38, hungshinlee\}@gmail.com}

\maketitle

\begin{abstract}
Children speech recognition is indispensable but challenging due to the diversity of children's speech. In this paper, we propose a filter-based discriminative autoencoder for acoustic modeling. To filter out the influence of various speaker types and pitches, auxiliary information of the speaker and pitch features is input into the encoder together with the acoustic features to generate phonetic embeddings. In the training phase, the decoder uses the auxiliary information and the phonetic embedding extracted by the encoder to reconstruct the input acoustic features. The autoencoder is trained by simultaneously minimizing the ASR loss and feature reconstruction error. The framework can make the phonetic embedding purer, resulting in more accurate senone (triphone-state) scores. Evaluated on the test set of the CMU Kids corpus, our system achieves a 7.8\% relative WER reduction compared to the baseline system. In the domain adaptation experiment, our system also outperforms the baseline system on the British-accent PF-STAR task.
\end{abstract}


\section{Introduction}
\label{sec:intro}

Today, the technology of automatic speech recognition (ASR) is mature enough to be applied to the daily life of adults. However, children need it but cannot benefit as much as adults. According to the research in \cite{Potamianos2003}, the word error rate (WER) of children speech recognition can reach 5 times that of adults. The first reason is the lack of children corpora. It is usually much easier to collect transcribed adult speech from news broadcasts and regular recordings. But the above scenarios are rare for children. By 2016, there were only 13 children speech corpora that contained partial or complete word transcriptions \cite{Chen2016}. To address the limitation caused by the lack of children resources, many efforts have been made to jointly use a large adult speech corpus and a relatively small children speech corpus for training acoustic models for children speech recognition \cite{Shahnawazuddin2016,Ahmad2017,Shahnawazuddin2017a,Shahnawazuddin2017b,Tong2017}. Experiments have shown that the joint training approach (a.k.a. multi-condition training) can reduce WER compared to the training method that uses children speech alone.

\begin{figure}[t!]
\centering
\includegraphics[width=0.47\textwidth]{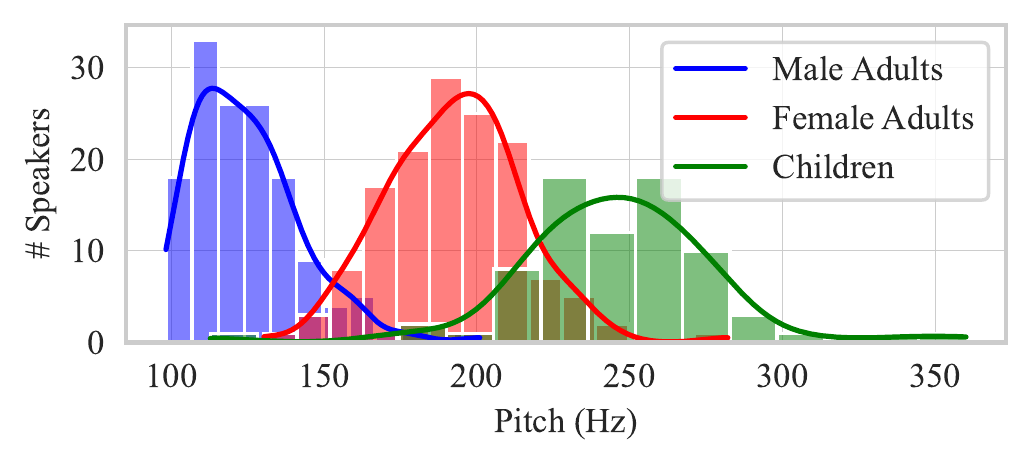}
\vspace{-5pt}
\caption{Kernel density estimate (KDE) plots of the pitch distributions in the CMU Kids and WSJ corpora. The horizontal axis represents the average pitch computed in every 25 Hz interval for each speaker. The numbers of male adults, female adults, and children are 142, 141 and 76, respectively.}
\label{fig:pitch_orig}
\vspace{-15pt}
\end{figure}

In addition to changes in volume, prosody, and articulation that make adult and children voices different, research has shown that the pitch of children speech is not only higher but has greater changes. This phenomenon indicates that the acoustic features of high-pitched speech are not sufficient to train children ASR, although the acoustic features of low-pitched speech can be supplemented from adult corpora.
Figure \ref{fig:pitch_orig} shows the pitch distribution of 283 adult speakers of the WSJ corpus and 76 child speakers of the CMU Kids corpus. The pitch value of a person in Hz is derived by averaging the estimated pitch values of a series of voiced frames in all utterances he/she spoke. Obviously, the distribution of pitch in the group of children is wider, twice that of male or female adults. This phenomenon indicates that the features derived from the high-pitched speech are insufficient for training children ASR, but the features from low-pitched speech can be supplemented by adult speech.

Moreover, the studies in \cite{Ghai2009,Shahnawazuddin2016} show that children's Mel-frequency cepstrum coefficients (MFCCs) are not immune to the effect of pitch, especially in the case of higher pitches. In \cite{Shahnawazuddin2016}, the algorithm for feature extraction was reformed to filter out the high-pitched components according to the speaker's pitch level. In \cite{Ahmad2017}, pitch scaling was used to adjust the pitch of the child downward, so that the pitch variation of the adjusted child's speech could correspond to the pitch range of the adult. In \cite{Shahnawazuddin2017a}, not only pitch adaptation was considered, but the children speaking rate was modified. In \cite{Serizel2014b}, vocal tract length normalization was performed to compensate the spectral variation caused by the difference in the vocal tract length among adults and children. The above methods focus on feature-level manipulation, with the purpose of reducing the inter-speaker acoustic variability.

As for the methods applied to the model itself, researchers have focused on adaptive \cite{Shivakumar2020,Tong2017} or multi-task learning \cite{Tong2017} of acoustic models. In \cite{Shivakumar2020}, the transfer learning strategies for adapting the acoustic and pronunciation variability were discussed separately. In \cite{Tong2017}, two different softmax layers were optimized with adult speech and children speech respectively to differentiate the phonemes of children speech from those of adult speech. Nonetheless, any additional adjustments to the model cannot avoid the risk of catastrophic forgetting or over-fitting due to the scarcity of the children corpus. To reduce the inter-speaker distinction, another potential approach is to provide auxiliary information, such as speaker embeddings (e.g., i-vectors and x-vectors) \cite{Peddinti2015a,Snyder2017,Snyder2018} and prosodic features (e.g., pitch and loudness) \cite{Kathania2018}, to the model. 

In \cite{Yang2017,Huang2019}, discriminative autoencoder-based (DcAE) acoustic modeling was proposed to separate the acoustic feature into the components of phoneme, speaker and environmental noise. Such model-space innovation takes a great advantage of unsupervised learning to extract pure phonetic components from the acoustic feature to better recognize speech. Inspired by this creativity, we combine the strength of the auxiliary information, i.e., the i-vector and pitch-related vector (called p-vector in this paper \cite{Ghahremani2014}) into autoencoder-based acoustic modeling to deal with very high-pitched speech (mainly children's speech). Because the use of i-vector and/or p-vector in DcAE-based acoustic modeling can be regarded as a filtering mechanism with inducers to purify the phonetic information in the acoustic feature, our model is called filter-based DcAE (f-DcAE for short).

\section{Proposed Model}
\label{sec:format}

\subsection{Filtering Mechanism for Acoustic Modeling}

Our ASR model belongs to the Gaussian mixture models (GMM)/deep neural networks (DNN)/hidden Markov models (HMM) topology and is developed based on the {\ttfamily chain} structure of the Kaldi toolkit. The basic DNN training process takes the MFCCs and i-vectors as input with two loss functions: cross-entropy (CE) for frame-level training and lattice free maximum mutual information (LF-MMI) \cite{Povey2016} for sentence-level training. The main function of using i-vectors for acoustic modeling is to eliminate factors such as speaker variation and channel mismatch in the acoustic features, thereby purifying the phonetic embedding. In this study, considering the extremely high-pitched characteristics of children, in addition to the i-vector, the p-vector is also used to reduce the impact of pitch on model adaptation. 

\begin{figure}[t!]
\centering
\includegraphics[width=0.49\textwidth]{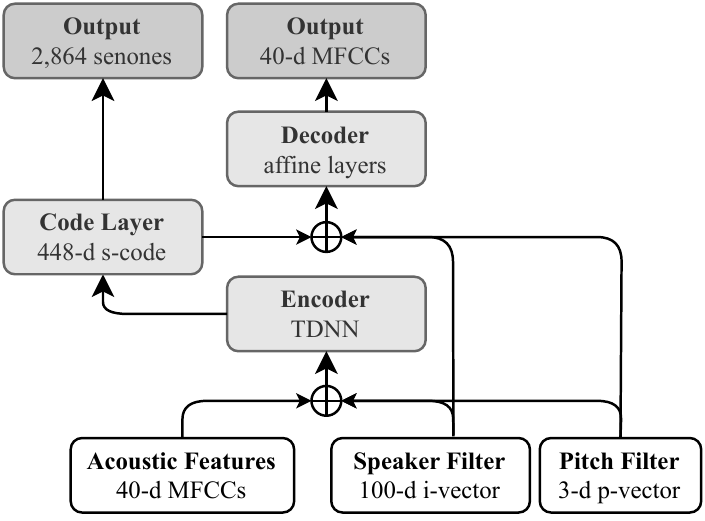}
\vspace{-15pt}
\caption{The structure of f-DcAE, where $\bigoplus$ denotes the concatenation operator, and the two outputs (black boxes) are associated with three kinds of objective functions. There is no need to use the decoder in the recognition phase.}
\label{fig:structure}
\vspace{-10pt}
\end{figure}

To import the advantage of unsupervised learning in \cite{Yang2017,Huang2019} into our acoustic modeling, the DNN model for generating the emission probabilities of the output labels can be regarded as the encoder, and the decoder is used to reconstruct the acoustic features, as shown in Figure \ref{fig:structure}. The output of the penultimate layer of the encoder is supposed to be a pure phoneme-related vector without any information irrelevant to the phonetic content. Therefore, it can be regarded as a latent embedding vector representing the phonetic information (called s-code below). The s-code is concatenated with the i-vector, the p-vector, or the fusion of the two as the input of the decoder to reconstruct the original MFCCs.

Generally speaking, the encoder functions to identifying senones, i.e., triphone-states. (There are a total of 2,864 senones in this study.) In our proposed model, the encoder functions further like a senone filter that uses speaker and pitch information as guidance to remove impurities that are not related to senones from the input acoustic features. The participation of the decoder is to enhance the training of the senones filter. The significance of feature reconstruction in the decoder lies in the amplification on purifying the phonetic components in the s-code. The similarity between the input acoustic features and the reconstructed features relies on the high-quality s-code and the strong representations of speaker and pitch characteristics, which can be treated as three orthogonal factors of a speech utterance. By introducing a decoder in the training phase, the encoder is further indirectly guided to yield the s-code that better interprets the senone information, while the decoder attempts to reconstruct the original feature from the s-code together with the i-vector and/or the p-vector. The proposed model is called filter-based discriminative autoencoder (f-DcAE). Note that f-DcAE can be implemented on top of various DNN-based acoustic models, such as the long short-term memory (LSTM) \cite{Peddinti2017}, multi-head attention-based networks \cite{Qin2019a}, transformers \cite{Wang2020}, and time-delay neural network (TDNN) \cite{Peddinti2015b} by considering them as the encoder. Although adding a decoder will increase the size of the entire model, only the encoder is used in the recognition phase. Therefore, the time complexity of f-DcAE in the recognition process is the same as that of the encoder-only counterpart.

\subsection{Objective Functions}

\subsubsection{Senones-aware CE and LF-MMI}

The first ASR-related objective function of the proposed f-DcAE model $\theta$ is the cross-entropy (CE) between the predicted senone (triphone-state) scores and ground truth defined as
\begin{equation}
F_{CE} = -\sum_{u=1}^{U} \sum_{t=1}^{T_u} \log p_{\theta}(\mathbf{s}_{ut}|\mathbf{x}_{ut}),
\label{eq:ce}
\end{equation}
where $U$ is the total number of training utterances, $T_u$ is the total number of frames of utterance $u$, and $\mathbf{x}_{ut}$ and $\mathbf{s}_{ut}$ are the $t$-th input frame and the corresponding senone label derived by GMM-based forced alignment, respectively. The second function relates to maximum mutual information (MMI), designed to maximize the probability of the reference transcription while minimizing the probability of all other transcriptions \cite{Vesely2013}. As in \cite{Hadian2018}, the LF-MMI is defined by,
\begin{equation}
F_{LF-MMI} = \sum_{u=1}^{U} \log \dfrac{p_{\theta}(\mathbf{x}_{u}|M_{\mathbf{w}_u})P(M_{\mathbf{w}_u})}{p_{\theta}(\mathbf{x}_{u})},
\label{eq:lf_mmi}
\end{equation}
where $\mathbf{w}_u$ is the reference word sequence of $u$, and the composite HMM graph $M_{\mathbf{w}_u}$ represents all the possible state sequences pertaining to $\mathbf{w}_u$, and is called the numerator graph. The denominator in Eq. (\ref{eq:lf_mmi}) can be further expressed as
\begin{equation}
p_{\theta}(\mathbf{x}_{u})={\sum_{\mathbf{w}} p_{\theta}(\mathbf{x}_{u}|M_\mathbf{w})P(M_\mathbf{w})}=p_{\theta}(\mathbf{x}_{u}|M_{den}),
\label{eq:lf_mmi_den}
\end{equation}
where $M_{den}$ is an HMM graph that includes all possible sequences of words, and is called the denominator graph. The denominator graph has traditionally been estimated using lattices. This is because the full denominator graph can become large and make the computation significantly slow. More recently, Povey \textit{et al.} derived MMI training of HMM-DNN models using a full denominator graph \cite{Povey2016}.

\subsubsection{Reconstruction errors and the final objective function}

The most straightforward strategy for evaluating the similarity between reconstructed features and input features is through the mean squared error (MSE) defined as
\begin{equation}
F_{MSE} = \sum_{u=1}^{U} \sum_{t=1}^{T_u} \left|\left| \mathbf{x}'_{ut}-\mathbf{x}_{ut}\right|\right|^{2}_{2},
\label{eq:mse}
\end{equation}
where $\mathbf{x}'_{ut}$ is the reconstructed feature vector of the input feature vector $\mathbf{x}_{ut}$, and $\left|\left|\cdot \right|\right|^{2}_{2}$ is the 2-norm operator. 

The final loss function $L$ to be minimized is the combination of the senone-aware CE, LF-MMI, and reconstruction error:
\begin{equation}
L = \alpha F_{CE} - F_{LF-MMI} + \beta F_{MSE},
\label{eq:final}
\end{equation}
where $\alpha$ and $\beta$ are the regularization penalty and weighting factor for $F_{CE}$ and $F_{MSE}$, respectively. We set $\alpha$ to $5$ followed by most recipes, and $\beta$ was heuristically determined based on the development set in our experiments.

\section{Experimental Settings}

\subsection{Datasets, Features, and Filters}

Children speech in our experiments came from the CMU Kids corpus (LDC97S63\footnote{\url{https://catalog.ldc.upenn.edu/LDC97S63}}), which contains 9.1 hours of speech from 151 children. The age range of children is 6--11 years old. The speech content mainly consists of reading sentences from sources such as storybooks or textbooks. The training set ({\ttfamily tr\_cmu}) contains 6.34 hours of speech from 76 kids, and the test set ({\ttfamily cmu}) contains 2.75 hours of speech from other 75 kids. The WSJ corpus was used as the adult speech corpus. The training set ({\ttfamily tr\_wsj}) contains 81.48 hours of speech from 283 adults, the first test set ({\ttfamily eval92}) contains 0.7 hour of speech, and the second test set ({\ttfamily dev93}) contains 1.08 hours of speech. We combined {\ttfamily tr\_cmu} and {\ttfamily tr\_wsj} as a joint training set for multi-condition training in the experiments. In addition to the American-accent CMU Kids corpus, the PF-STAR British English children speech corpus \cite{Batliner2005} was also used in the experiments. The age range of children in PF-STAR is 4--14 years old. It contains 14.2 hours of speech from 152 children. The training set ({\ttfamily tr\_pfstar}) contains 8.4 hours of speech from 92 kids, and the test set ({\ttfamily pfstar}) contains 5.82 hours of speech from other 60 kids.

\begin{figure}[t!]
\centering
\includegraphics[width=0.48\textwidth]{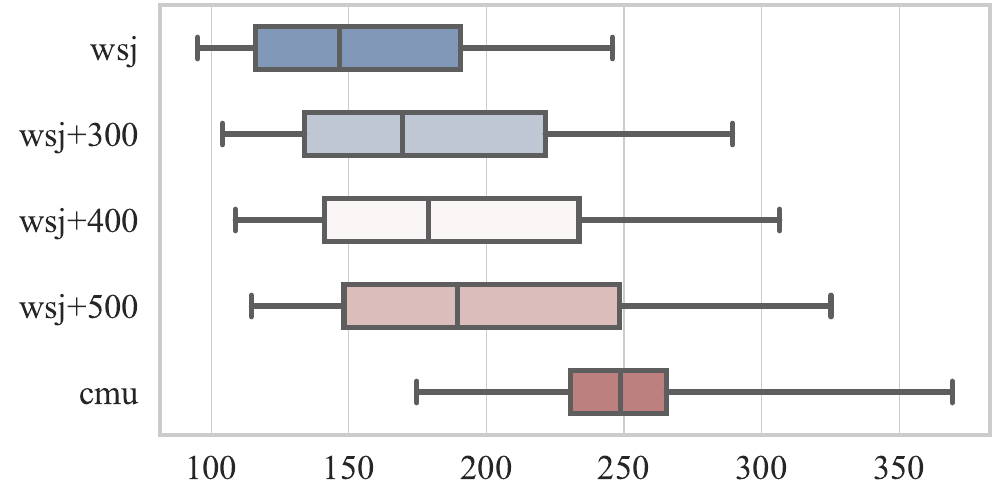}
\vspace{-10pt}
\caption{Box plots of the pitch distributions in Hertz of the simulated children's speech from {\ttfamily eval92} and {\ttfamily dev93} (denoted by {\ttfamily wsj+xxx}) and the test set of the CMU Kids corpus (denoted by {\ttfamily cmu}).}
\label{fig:pitch_test}
\vspace{-10pt}
\end{figure}

To evaluate the ability of the acoustic model to filter out the effects of pitch, we intentionally created the high-pitched versions of {\ttfamily eval92} and {\ttfamily dev93} to simulate children speech. We used Sound eXchange (SoX\footnote{\url{http://sox.sourceforge.net/sox.html}}) to raise the pitch of adult speech by setting the parameter {\ttfamily pitch} to {\ttfamily +300}, {\ttfamily +400}, and {\ttfamily +500} without changing the speaking rate, where the units of these values are cents, not hertz. Figure \ref{fig:pitch_test} shows the pitch distribution of the original test sets and simulated versions of {\ttfamily eval92} and {\ttfamily dev93}. We found that once the pitch was set to {\ttfamily +600} or beyond, the adjusted speech sounded considerably unnatural, not similar to the speech of ordinary children and adults. In addition, the simulated speech was occasionally mixed with sharp mechanical noises, which could hardly represented children speech. Therefore, we used {\ttfamily +300}, {\ttfamily +400}, and {\ttfamily +500} to generate the simulated children test sets.

The size of training data was tripled using speed and volume perturbation. The 40-dimensional high-resolution raw MFCCs were used as the acoustic features.

The i-vector was 100-dimensional and was extracted every 10 frames \cite{Dehak2011}. We computed the average of pitch, delta-pitch and the normalized cross correlation function every 10 frames to derive a 3-dimensional p-vector \cite{Ghahremani2014}.

\begin{table*}
\centering
\caption{WERs (\%) and their relative reductions (\%) with respect to various test sets.}
\vspace{-5pt}
\label{tab:results}
\centering
\begin{tabular*}{\linewidth}{@{\extracolsep{\fill}} cccccccc}
\toprule
\multicolumn{1}{c}{\textbf{Filters}} &
\multicolumn{1}{c}{\textbf{None}} &
\multicolumn{2}{c}{\textbf{i-vector}} &
\multicolumn{2}{c}{\textbf{p-vector}} &
\multicolumn{2}{c}{\textbf{i-vector $\bigoplus$ p-vector}} \\
\cmidrule{1-1}
\cmidrule{2-2}
\cmidrule{3-4}
\cmidrule{5-6}
\cmidrule{7-8}
\multicolumn{1}{c}{\textbf{Test Set\textbackslash Model}} & 
\textbf{Baseline(o)} &
\textbf{Baseline(i)} & 
\textbf{f-DcAE(i)} &
\textbf{Baseline(p)} &
\textbf{f-DcAE(p)} &
\textbf{Baseline(i+p)} &
\textbf{f-DcAE(i+p)} \\
\midrule
\midrule
\multicolumn{1}{c}{{\ttfamily cmu}} & 18.82 & 17.73 & \textbf{16.82} (5.13) & 18.92 & \textbf{17.45} (7.77) & 17.58 & \textbf{16.44} (6.48) \\
\midrule
\multicolumn{1}{c}{{\ttfamily eval92+300}} & 2.96 & 2.68 & \textbf{2.60} (3.85) & 2.98 & \textbf{2.55} (14.43) & 2.89 & 2.89 (0.00) \\
\multicolumn{1}{c}{{\ttfamily eval92+400}} & 3.81 & 3.51 & \textbf{3.35} (4.56) & \textbf{3.79} & 3.95 (-4.22) & 3.83 & \textbf{3.33} (13.05) \\
\multicolumn{1}{c}{{\ttfamily eval92+500}} & 5.64 & 5.30 & \textbf{5.00} (5.66) & 5.37 & \textbf{5.07} (5.59) & 5.46 & \textbf{5.23} (4.21)\\
\midrule
\multicolumn{1}{c}{{\ttfamily dev93+300}} & 5.88 & 5.40 & \textbf{5.10} (5.56) & \textbf{5.67} & 5.71 (-0.71) & 5.42 & \textbf{5.33} (1.67) \\
\multicolumn{1}{c}{{\ttfamily dev93+400}} & 6.87 & 6.73 & \textbf{6.47} (3.87) & 6.74 & \textbf{6.45} (4.30) & 6.59 & \textbf{6.27} (4.86) \\
\multicolumn{1}{c}{{\ttfamily dev93+500}} & 9.27 & 9.62 & \textbf{8.78} (8.73) & 8.93 & \textbf{8.77} (1.79) & 9.56 & \textbf{8.89} (7.00) \\
\bottomrule
\end{tabular*}
\vspace{-10pt}
\end{table*}

\subsection{Model Configuration}

We implemented our baseline model based on TDNN \cite{Peddinti2015b}, which has been used for children's ASR in \cite{Wu2019}, under the {\ttfamily chain} setting of the Kaldi toolkit. The TDNN model was composed of one affine layer, eight 448-dimensional TDNN layers, and one 448-dimensional dense layer before the softmax function. It was trained by minimizing the loss function consisting of CE and LF-MMI with the L2 regularization. We applied our training data (i.e., {\ttfamily tr\_wsj} + {\ttfamily tr\_cmu}) to proceed with basic recipes, including feature extraction, GMM training/alignment, and DNN training.

To realize our f-DcAE model in Figure \ref{fig:structure}, on top of the baseline TDNN model, we added a decoder, which consists of four affine layers, each with 128 nodes. The first affine layer of the decoder was fed with the phonetic embedding (i.e., s-code, the last TDNN layer of the baseline model) extracted by the encoder (the first seven TDNN layers of the baseline model) as well as the i-vector and/or the p-vector.

The 4-gram language model (LM) and an enhanced lexicon, same as those in \cite{Peddinti2015b}, were used. The perplexities for {\ttfamily cmu}, {\ttfamily pfstar}, {\ttfamily eval92} and {\ttfamily dev93} are 529.7, 693.2, 164.9, and 200.4, respectively. The ratios of the numbers of out-of-vocabulary words to total words for {\ttfamily cmu}, {\ttfamily pfstar}, {\ttfamily eval92} and {\ttfamily dev93} are 0/12180, 61/24838, 2/5700, and 17/8334, respectively. The perplexities of {\ttfamily cmu} and {\ttfamily pfstar} stand out dramatically because the LM is trained with news transcripts, and {\ttfamily eval92} and {\ttfamily dev93} are in the same domain. Hence, there is a domain mismatch between the LM and the two children test sets. We did not use the training transcriptions of the two children's corpora for LM training, because we found that their training and test transcriptions highly overlap each other.

\begin{table}[t!]
\caption{WERs (\%) of {\ttfamily cmu} and {\ttfamily pfstar} with one-epoch adaptation using {\ttfamily tr\_cmu} and {\ttfamily tr\_pfstar}, respectively.}
\vspace{-5pt}
\label{tab:results2}
\begin{tabular*}{\linewidth} {@{\extracolsep{\fill}} ccccc}
\toprule
\multicolumn{1}{c}{\textbf{Adapt./Test Sets}} &
\multicolumn{2}{c}{{\ttfamily tr\_cmu/cmu}} &
\multicolumn{2}{c}{{\ttfamily tr\_pfstar/pfstar}} \\
\cmidrule{1-1}
\cmidrule{2-3}
\cmidrule{4-5}
\multicolumn{1}{c}{\textbf{One-epoch Adapt.}} &
\multicolumn{1}{c}{\textbf{no}} &
\multicolumn{1}{c}{\textbf{yes}} &
\multicolumn{1}{c}{\textbf{no}} &
\multicolumn{1}{c}{\textbf{yes}}\\
\midrule
\midrule
\multicolumn{1}{c}{Baseline(i)} & 17.73 & 15.41 & 71.21 & 19.28 \\
\multicolumn{1}{c}{f-DcAE(i)} & 16.82 & \textbf{14.60} & 72.03 & \textbf{18.77} \\
\bottomrule
\end{tabular*}
\vspace{-10pt}
\end{table}

\section{Experimental Results and Discussion}
\label{sec:typestyle}

\subsection{TDNN-based Baseline versus f-DcAE}
Table \ref{tab:results} shows the WERs of the TDNN-based baseline and f-DcAE evaluated on different test sets. The model, consisting of a TDNN-based encoder with i-vector but no decoder, is regarded as a baseline that has been widely used in many studies. This baseline is denoted as Baseline(i) in Table \ref{tab:results}. Baseline(p) denotes the baseline model using p-vector instead of i-vector. Baseline(i+p) denotes the baseline model that uses both i-vector and p-vector. Baseline(o) denotes the baseline model without using any additional embedding. Corresponding to baseline models Baseline(i), Baseline(p), and Baseline(i+p), the proposed models with a decoder are denoted as f-DcAE(i), f-DcAE(p), and f-DcAE(i+p).

From Table \ref{tab:results}, we can see that f-DcAE(i), f-DcAE(p), and f-DcAE(i+p) are always better than Baseline(i), Baseline(p), and Baseline(i+p), respectively, on the {\ttfamily cmu} test set. The WER is relatively reduced by 5.1\%, 7.8\%, and 6.5\%, respectively. This result shows that no matter which type of additional embedding is used, the proposed DcAE-based filtering mechanism can improve the recognition performance. As for the results of the simulated test sets, we can see that in most cases, f-DcAE(i), f-DcAE(p), and f-DcAE(i+p) are better than Baseline(i), Baseline(p), and Baseline(i+p), respectively. In summary, these results generally confirm the effectiveness of the proposed DcAE-based filtering mechanism for acoustic modeling, especially for the challenging children speech recognition task. Next, we focus on the results of the simulated (pitch-shifted) test sets. From Table \ref{tab:results}, we can see that the WER has a clear increasing trend with the rise of {\ttfamily pitch} shift (from {\ttfamily +300} to {\ttfamily +500}). The proposed model with the filtering mechanism can achieve more WER reduction on the test set with more pitch shifts, e.g., f-DcAE(i+p) reduces WER by 7.00\% on the {\ttfamily dev93+500} test set compared to Baseline(i+p).

\begin{table}[t!]
\caption{WERs (\%) of {\ttfamily cmu} and {\ttfamily pfstar} with training from scratch using {\ttfamily tr\_cmu\textsuperscript{*}} ({\ttfamily tr\_wsj}+{\ttfamily tr\_cmu}) and {\ttfamily tr\_pfstar\textsuperscript{*}} ({\ttfamily tr\_wsj}+{\ttfamily tr\_pfstar}), respectively.}
\vspace{-5pt}
\label{tab:results3}
\begin{tabular*}{\linewidth} {@{\extracolsep{\fill}} ccc}
\toprule
\multicolumn{1}{c}{\textbf{Training/Test Sets}} &
\multicolumn{1}{c}{{\ttfamily tr\_cmu\textsuperscript{*}/cmu}} &
\multicolumn{1}{c}{\ttfamily tr\_pfstar\textsuperscript{*}/pfstar} \\
\midrule
\midrule
\multicolumn{1}{c}{Baseline(i)} & 17.73 & 20.57 \\
\multicolumn{1}{c}{f-DcAE(i)} & \textbf{16.82} & \textbf{19.39} \\
\bottomrule
\end{tabular*}
\vspace{-10pt}
\end{table}

\subsection{Adaptation with In-domain and Out-of-domain Data}
In this experiment, we studied the effect of model adaptation. We compared Baseline(i) and f-DcAE(i). The models in Table \ref{tab:results} were used as the seed models and were adapted with one-epoch training by {\ttfamily tr\_cmu} (in-domain data) and {\ttfamily tr\_pfstar} (out-of-domain data). The results are shown in Table \ref{tab:results2}. We can see that adapted with one-epoch training by {\ttfamily tr\_cmu}, Baseline(i) can reduce the WER by 13.09\% (from 17.73\% to 15.41\%), while f-DcAE(i) can reduce the WER by 13.20\% (from 16.82\% to 14.60\%), when evaluated on {\ttfamily cmu}. Even though their relative WER reductions are comparable, the WER of f-DcAE(i) is lower than that of Baseline(i) (14.60\% vs 15.41\%). When evaluated on {\ttfamily pfstar}, we can see that before adaptation, both initial models performed poorly, because {\ttfamily pfstar} is in British accent while the training data {\ttfamily tr\_wsj+tr\_cmu} is in American accent. With one-epoch adaptation by {\ttfamily tr\_pfstar}, the WERs of both models were significantly reduced, and f-DcAE(i) achieved a lower WER than Baseline(i) (18.77\% vs 19.28\%).

\subsection{Training from scratch for PF-STAR}
In this experiment, we investigated whether our proposed model can work well under a specific multi-condition training condition, where the training data contain the speech of American adults and British children. The models were trained from scratch using {\ttfamily tr\_wsj}+{\ttfamily tr\_cmu} (both in American accent) or {\ttfamily tr\_wsj}+{\ttfamily tr\_pfstar} (the former in American accent and the latter in British accent) with sufficient training epochs. The results are shown in Table \ref{tab:results3}. For the {\ttfamily cmu} task, f-DcAE(i) can reduce the WER by 5.13\% compared to Baseline(i) (from 17.73\% to 16.82\%). For the {\ttfamily pfstar} task, f-DcAE(i) can reduce the WER by 5.74\% compared to Baseline(i) (from 20.57\% to 19.39\%). The result confirmed that an adult training corpus in another accent can help improve the performance of children ASR. In addition, the result once again confirmed that the proposed f-DcAE model is better than the baseline model.

\section{Conclusions and Future Work}
\label{sec:majhead}

In this paper, we have proposed a filter-based discriminative autoencoder (f-DcAE) architecture for acoustic modeling. Both the encoder and decoder are fed with the speaker and/or pitch filters. With the help of auxiliary information, the encoder purifies the phonetic information in the input acoustic feature, while the decoder reconstructs the input acoustic feature from the phonetic information extracted by the encoder. Experimental results have shown that our f-DcAE models outperform the counterpart baseline models without using the autoencoder architecture and the filtering mechanism.

In future work, a more non-experimental proof that the code layer (s-code) contains less speaker and pitch information in f-DcAE than the baseline model will be provided. For example, we will try to project each kind of s-codes into a plane using t-SNE \cite{Maaten2008} to visualize and compare their degrees of separation for speaker or pitch classes. Moreover, we will study another decent feature of f-DcAE, unsupervised pre-training without ASR losses, so that we can make the best use of a large amount of unlabeled children speech for further domain adaptation.

\bibliographystyle{IEEEtran}
\bibliography{references.bib}
\end{document}